\documentclass[journal]{IEEEtran}

\usepackage{newtxtext,newtxmath}
\usepackage{amsmath,amsfonts}
\usepackage{algorithm}
\usepackage{array}
\usepackage[caption=false,font=normalsize,labelfont=sf,textfont=sf]{subfig}
\usepackage{textcomp}
\usepackage{stfloats}
\usepackage{url}
\usepackage{verbatim}
\usepackage{graphicx}
\usepackage{xcolor}
\usepackage{multirow}
\usepackage{booktabs}
\usepackage{stfloats}
\usepackage[capitalize]{cleveref}

\newcommand{\ie}{\textit{i}.\textit{e}.}

\definecolor{deepgreen}{RGB}{50, 235, 50}
\usepackage{algorithmicx}
\usepackage{algpseudocode}
\usepackage{tikz}
\usepackage{pgfplots}

\begin{document}
\title{Vision Foundation Model Driven Foreground-Aware Pseudo-LiDAR Generation for Monocular 3D Object Detection}

\author{Bonan Ding,
	Jin Xie,
	Jing Nie,
	Jiale Cao,
	and Yanwei Pang}
\maketitle
\begin{abstract}
Pseudo-LiDAR has become a promising paradigm for monocular 3D object detection by transforming monocular images into point cloud representations that can be processed by LiDAR-based 3D object detectors. Recent vision foundation models provide powerful geometric and semantic priors, creating new opportunities for improving the quality of pseudo-LiDAR generation. However, effectively exploiting these priors to produce reliable pseudo-LiDAR remains challenging due to inaccurate depth estimation, insufficient foreground awareness, and redundant background points.
In this paper, we propose VFMM3D, a vision foundation model-driven framework for foreground-aware pseudo-LiDAR generation. VFMM3D leverages the depth prior provided by the Depth Anything Model (DAM) and the foreground prior provided by the Segment Anything Model (SAM). A foreground-aware pseudo-LiDAR painting operation is introduced to associate projected 3D points with object-level foreground information, thereby strengthening object structures and suppressing irrelevant background regions. Furthermore, a sparsification strategy is developed to remove redundant points, reduce computational overhead, and improve the compatibility of the generated point clouds with LiDAR-based 3D object detectors.
Comprehensive experiments are conducted on two challenging 3D object detection datasets, KITTI and Waymo. Our VFMM3D establishes a new state-of-the-art performance on both datasets. Additionally, experimental results demonstrate the generality of VFMM3D, showcasing its seamless integration into various LiDAR-based 3D object detectors.
\end{abstract}

\section{Introduction}

Monocular 3D object detection aims to estimate the locations, dimensions, and orientations of objects in 3D space from a single RGB image. 
The rise of autonomous systems~\cite{survey2025} has brought about an era where perceiving and understanding~\cite{multiunderstanding} the environment in three dimensions is not just advantageous but imperative. 
Compared with LiDAR-based perception systems, monocular cameras provide a low-cost and widely available sensing solution, making monocular 3D object detection an attractive research direction for autonomous driving, advanced driver-assistance systems (ADAS), robotics, and other applications. 
However, extracting reliable 3D information from a single 2D image remains challenging due to the inherent ambiguity caused by perspective projection.

Existing monocular 3D object detection approaches rely on geometry-based methods~\cite{MonoPair, MonoPSR, Unicuboid} or depth estimation techniques~\cite{CaDNN, MonoGround, DDCDC}.
However, accurately recovering reliable 3D geometry from monocular images remains challenging, particularly in complex and dynamic scenes. 
To alleviate this limitation, pseudo-LiDAR has emerged as a promising paradigm for monocular 3D object detection by transforming monocular images into point cloud representations and leveraging LiDAR-based 3D detectors for object detection. 
By bridging the gap between image-based perception and point cloud-based detection, pseudo-LiDAR enables the utilization of well-established LiDAR detection architectures. 
Nevertheless, the effectiveness of pseudo-LiDAR heavily depends on the quality of the generated point clouds, where depth estimation errors, redundant background points, and insufficient foreground structures may degrade the final detection performance.

Recent vision foundation models provide new opportunities for improving pseudo-LiDAR generation by offering powerful visual priors learned from large-scale and diverse datasets. 
The Depth Anything Model (DAM)~\cite{DAM} provides strong monocular depth estimation capability, offering valuable geometric priors for recovering 3D structures from images. 
Meanwhile, the Segment Anything Model (SAM)~\cite{SAM} provides class-agnostic object masks with accurate spatial boundaries, offering foreground priors for distinguishing object regions from background content. 
These priors are highly relevant to pseudo-LiDAR generation, where depth information determines the spatial locations of projected points and foreground information helps preserve object-related structures. 
However, the outputs of DAM and SAM remain in the image domain and cannot be directly utilized by LiDAR-based 3D detectors. 
Therefore, effectively transforming vision foundation model priors into structured, foreground-aware, and detector-compatible pseudo-LiDAR representations remains an important yet underexplored problem.

To address the aforementioned challenge, we propose VFMM3D, a vision foundation model-driven framework for generating foreground-aware pseudo-LiDAR representations for monocular 3D object detection. 
VFMM3D exploits the depth prior from the Depth Anything Model (DAM) and the foreground prior from the Segment Anything Model (SAM) to convert monocular images into structured 3D point cloud representations. 
Specifically, a foreground-aware pseudo-LiDAR painting operation is introduced to associate projected 3D points with object-level foreground information, enhancing object structures while suppressing irrelevant background regions. 
Furthermore, a sparsification strategy is developed to remove redundant points and improve the compatibility between generated pseudo-LiDAR and LiDAR-based 3D object detectors. 
The generated pseudo-LiDAR representations can be directly processed by various off-the-shelf LiDAR-based 3D detectors without modifying their detection architectures.

In conclusion, the main contributions of this work are summarized as follows:

\begin{itemize}
    \item We propose VFMM3D, a vision foundation model-driven framework for foreground-aware pseudo-LiDAR generation in monocular 3D object detection. By exploiting depth and foreground priors from DAM and SAM without task-specific fine-tuning of these foundation models, VFMM3D generates structured pseudo-LiDAR representations that can be processed by various LiDAR-based 3D detectors.

    \item We introduce a foreground-aware pseudo-LiDAR painting operation that effectively integrates depth and foreground priors in 3D space, enhancing object-related structures while suppressing irrelevant background regions.

    \item We develop a sparsification strategy to reduce redundant pseudo-LiDAR points and computational overhead while preserving informative geometric structures, improving the compatibility between generated representations and LiDAR-based 3D detectors.

    \item Extensive experiments on the KITTI and Waymo datasets demonstrate that VFMM3D achieves superior performance compared with existing monocular 3D object detection methods. Additional evaluations with multiple LiDAR-based 3D detectors verify the effectiveness, generality, and detector compatibility of the proposed framework.
\end{itemize}

\section{Related Work}
\textbf{Vision Foundation Models.}
Vision Foundation Models (VFMs) represent a significant advancement in the realm of computer vision, offering versatile solutions across various tasks owing to their robust pre-training on extensive datasets. Among VFMs, Vision Transformers (ViTs)\cite{ViT} stand as pivotal models, trained on colossal datasets such as LVD-142M\cite{Dinov2}. The efficacy of ViT is further amplified through approaches like DINO~\cite{Dino} and DINOv2~\cite{Dinov2}, which leverage self-supervised learning coupled with knowledge distillation techniques. 
One notable application of VFMs is evident in the Segment Anything Model (SAM)~\cite{SAM}, which is particularly designed to be adept at generating precise masks for individual elements within images, showcasing its ability to handle detailed object segmentation. {It was trained} on a vast dataset SA-1B encompassing 11 million images and 1.1 billion masks. 
Similarly, the Depth Anything Model (DAM)~\cite{DAM} emerges as a robust solution for monocular depth estimation, enabling accurate projection of 2D imagery into 3D space. This capability is instrumental in providing depth information for each pixel in an image, facilitating a more comprehensive understanding of scene geometry. 
The integration of SAM and DAM within the VFMM3D model represents a novel approach to enhancing monocular 3D object detection. SAM's segmentation abilities are harnessed to refine foreground information within pseudo-LiDAR data, while DAM's depth estimation capabilities aid in projecting 2D images into 3D space, thereby extracting as much valuable information for 3D spatial representation as possible from 2D images. 

\noindent\textbf{Monocular 3D Object Detection.}
In the field of monocular 3D object detection, a variety of methods have been developed to extract 3D information from a single 2D image. 
MonoPair~\cite{MonoPair} is a typical geometry-based method that uses spatial relationships between objects to estimate 3D properties.  MonoFlex~\cite{MonoFlex} proposes a flexible framework that separates truncated objects and amalgamates multiple depth estimation approaches, including direct regression and geometric solutions from keypoints. MonoDLE~\cite{MonoDLE} and PGD~\cite{PGD} apply deep learning to directly predict 3D bounding boxes or depth on instances from images. 
Other geometry-centered approaches recover 3D attributes through explicit
image-to-world constraints. MCK-NET combines multi-instance depth with corner-based
geometric reasoning~\cite{MCK-NET}, while Li and Zhao model objects as dual
quadrics and recover their spatial parameters through geometric optimization
~\cite{DualQuadric}.
MonoRUn~\cite{MonoRUn} leverages self-supervised learning to establish dense correspondences and geometric relationships by using 3D bounding box annotations directly.
MonoDTR~\cite{MonoDTR} and MonoDETR~\cite{MonoDETR} introduce adaptive feature aggregation via a depth-guided transformer for monocular 3D object detection.
MonoPoly~\cite{MonoPoly} proposes an anchor-free keypoint-based monocular 3D detection family with a PolyPAN feature aggregation neck and attention-aware loss.
Pseudo-LiDAR-based methods~\cite{DD3D, Pseudo3d} generate a pseudo-LiDAR point cloud from the images and then process it using voxel-based techniques.
CaDDN~\cite{CaDNN} constructs bird’s-eye-view (BEV) representations by learning categorical depth distributions for each pixel. Then it recovers bounding boxes from the BEV projection.
MonoPSR~\cite{MonoPSR} estimates instance point clouds and refines proposals by enforcing alignment between object appearance and the projected point cloud.
M3D-RPN~\cite{M3D-RPN} introduces depth-aware convolutional layers within a 3D region proposal network to enhance the extracted features. 

\begin{figure*}[!t]
	\centering
	\includegraphics[width=\textwidth]{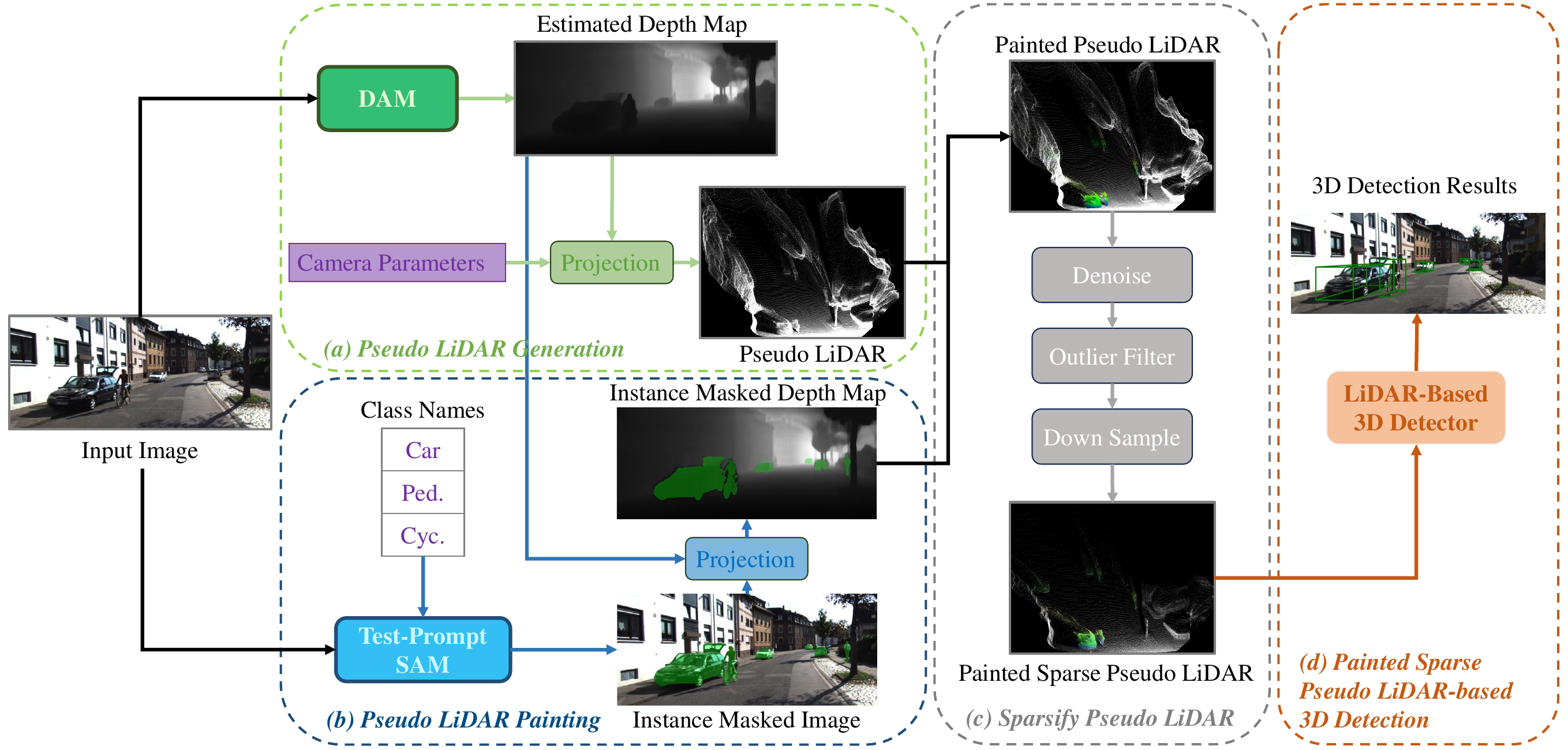}
	\caption{The overall architecture of VFMM3D. Our model consists of four parts: (a) Pseudo-LiDAR generation by DAM, (b) Pseudo-LiDAR painting by text-prompt SAM, (c) Pseudo-LiDAR sparsification, (d) 3D object detection by LiDAR-based detectors.}
	\label{fig:overview}
\end{figure*}

\noindent\textbf{LiDAR-based 3D Object Detection.}
Within the domain of LiDAR-based 3D object detection, there are two primary paradigms: point-based and voxel-based models. 
PointRCNN~\cite{Pointrcnn} exemplifies the former, harnessing PointNet++~\cite{Pointnet++}, the pioneer of point-based methods, for feature extraction from raw LiDAR point clouds. Contrarily, VoxelNet~\cite{Voxelnet} showcases a robust voxel feature encoding layer, transforming 3D LiDAR point clouds into equidistantly spaced 3D voxels. Building upon this voxel-centric approach, SECOND~\cite{SECOND} introduces a novel sparse 3D convolution layer tailored for LiDAR point clouds. Departing from traditional voxel methods, PointPillars~\cite{Pointpillars} employs pillars for feature encoding, enhancing network efficiency. Similarly, Voxel-RCNN~\cite{VoxelRCNN} proposes voxel query and voxel RoI pooling methods to fully exploit voxel features in a two-stage framework. Embracing a hybrid strategy, PV-RCNN~\cite{PVRCNN} integrates point and voxel advantages, employing voxel set abstraction to fuse multi-scale voxel features into sampled key points.

\section{Method}

\subsection{Task Definition}
The goal of monocular 3D object detection is to classify and localize objects of interest in 3D space by giving a single-view RGB image and its corresponding camera parameters. Each object is represented by its category score and bounding box (BBox), which is parameterized by its center coordinates $(x, y, z)$, size $(h, w, l)$, and orientation $\theta$.

\subsection{Framework Overview}
\label{subsec:overview}

Existing pseudo-LiDAR approaches mainly rely on monocular depth estimation to project image pixels into 3D space. However, depth-only projection may introduce inaccurate geometric structures and numerous redundant background points, resulting in noisy pseudo-LiDAR representations and degraded detection performance. To address these limitations, we introduce VFMM3D, a vision foundation model-driven framework that exploits depth and foreground priors from the Depth Anything Model (DAM) and Segment Anything Model (SAM) for foreground-aware pseudo-LiDAR generation.

Specifically, DAM is employed to estimate dense monocular depth maps, which provide geometric information for 3D point generation. Since depth estimation alone cannot distinguish object regions from background content, SAM is further utilized to extract foreground object regions. By integrating these two types of priors through the proposed pseudo-LiDAR painting operation, VFMM3D generates structured point cloud representations with enhanced object-related information and reduced background interference.

Furthermore, directly generated pseudo-LiDAR is typically denser than real LiDAR observations and contains redundant points, increasing computational cost and affecting detector performance. To improve efficiency and compatibility with existing LiDAR-based 3D object detectors, we propose a sparsification strategy to remove redundant points while preserving informative geometric structures. The whole pipeline of VFMM3D is shown in Fig.~\ref{fig:overview} and Algorithm~\ref{alg:overview}, which mainly contains four steps:

\begin{algorithm}[t]
	\caption{VFMM3D}\label{alg:overview}
	\begin{algorithmic}
		\Require{~\\
			RGB image $I \in \mathbb{R} ^ {H \times W \times 3}$.\\
		}
		\Ensure{~\\
			Predicted 3D bounding boxes $B_{3D} = \{(x_i, y_i, z_i, h_i, w_i, l_i)\}^M_{i=1}$
			\\}
	\end{algorithmic}
	\begin{algorithmic}
		\State $D = Depth~Generation(I)$ \Comment{Depth Map $D \in \mathbb{R} ^ {H \times W \times 1}$}
		\State $P = Projection(D)$ \Comment{Pseudo-LiDAR $P \in \mathbb{R} ^ {N \times 3}$}
		\State $I_M = Mask~Generation(I, B)$ \Comment{Instance Mask $I_M \in bool ^ {H \times W}$}
		\State $\hat{P} = Painting(D, I_M, P)$ \Comment{Painted Pseudo-LiDAR $\hat{P} \in \mathbb{R} ^ {N \times 6}$}
		\State $\bar{P} = Sparsification(P)$ \\ \Comment{Sparse Painted Pseudo-LiDAR $\bar{P} \in \mathbb{R} ^ {\bar{N} \times 6}$}
		\State $B_{3D} = LiDAR$-$Based~3D~Detector(\bar{P})$\\
	\end{algorithmic}
\end{algorithm}

\begin{enumerate}
	\item \textbf{Pseudo-LiDAR Generation}. At this step, we begin by acquiring the depth map using the DAM. Next, we project this depth map into 3D space and obtain pseudo-LiDAR data.
	\item \textbf{Pseudo-LiDAR Painting}. 
	
	SAM takes the RGB image as the input and combines text prompts to perform foreground object segmentation. The foreground object segmentation results are mapped to the depth map obtained by the DAM, which is employed to highlight the foreground object depth map and filter noises. Then the corresponding pseudo-LiDAR is also more accurate.
	
	\item \textbf{Pseudo-LiDAR Sparsification}. Since the number of pseudo-LiDAR points obtained from the depth map is much larger than the LiDAR point cloud in the real scene, we adopt a sparsification step to appropriately {sparsify} the painted pseudo-LiDAR to adapt to the final LiDAR-based {3D} object detector.
	\item \textbf{LiDAR-based 3D Detection.} To perform 3D object detection using the painted pseudo-LiDAR, we employ a LiDAR-based 3D object detector. This detector takes the painted pseudo-LiDAR generated from the previous steps as input and produces the final detection results.
\end{enumerate}

\subsection{Pseudo-LiDAR Generation with DAM}
\label{subsec:dam}

\textbf{Preliminaries.} The foundational step in our monocular 3D object detection pipeline is the estimation of depth from single-view images. We introduce the \textbf{Depth Anything Model (DAM)}~\cite{DAM} for robust monocular depth estimation (MDE).  DAM does not rely on technical modules but instead focuses on scaling up the dataset with a novel data engine designed to automatically collect and annotate approximately 62 million unlabeled images from various public large-scale datasets. The extensive use of unlabeled data significantly broadens the data coverage, thereby reducing the generalization error and improving the model's ability to handle diverse and challenging scenes.

The DAM leverages a two-pronged strategy to enhance its performance. First, it creates a more challenging optimization target through the use of data augmentation tools, compelling the model to actively seek additional visual knowledge and acquire robust representations. Second, it incorporates an auxiliary supervision mechanism that enforces the model to inherit rich semantic priors from pre-trained encoders. The above-mentioned approach not only enhances the model's performance in depth estimation but also provides a multi-task encoder capable of handling both middle-level and high-level perception tasks.

\noindent\textbf{Depth Estimation by DAM.} DAM is further fine-tuned with metric depth information from standard datasets such as NYUv2 and KITTI, setting new state-of-the-art records in metric depth estimation accuracy. Given a single image $I\in \mathbb{R}^{H\times W\times 3}$, we can get a reliable depth map $D\in \mathbb{R}^{H\times W\times 1}$ generated by the DAM. After getting the depth map, we then project it to {3D} space to get the pseudo-LiDAR in real-world coordinate $P=\{(x^{(n)},y^{(n)},z^{(n)})\}^N_{n=1}$ by the first project the depth map to the camera coordinate system:
\begin{equation}
	\begin{cases}
		z_c = d,\\
		x_c = \frac{(u - C_x) \times z}{f},\\
		y_c = \frac{(v - C_y) \times z}{f},
	\end{cases}
\end{equation}
where $d$ is the estimated depth of pixel $(u,v)$ in the depth map, and $(C_x, C_y)$ is the principal point of camera. $f_x$ and $f_y$ {are} the camera focal {lengths} along $x$ and $y$ axes, respectively. $N$ is the number of pixels. 
Then given the camera extrinsic matrix $M_E$:
\begin{equation}
	M_E = 
	\begin{bmatrix}
		R & t\\
		0^T & 1
	\end{bmatrix},
\end{equation}
where $M_E$ is a $4 \times 4$ matrix. $R$ is a $3 \times 3$ orthogonal unit matrix, also known as a rotation matrix. $t$ is a three-dimensional translation vector. 
Then we can get pseudo-LiDAR in the world coordinate system by:
\begin{equation}
	\begin{bmatrix}
		x\\
		y\\
		z\\
	\end{bmatrix}
	= M_E^{-1}
	\begin{bmatrix}
		x_c\\
		y_c\\
		z_c
	\end{bmatrix}
\end{equation}
\\
\subsection{Pseudo-LiDAR Painting with Text-Prompt SAM}
\label{subsec:sam}

\begin{figure}[t]
	\centering
	\includegraphics[width=\linewidth]{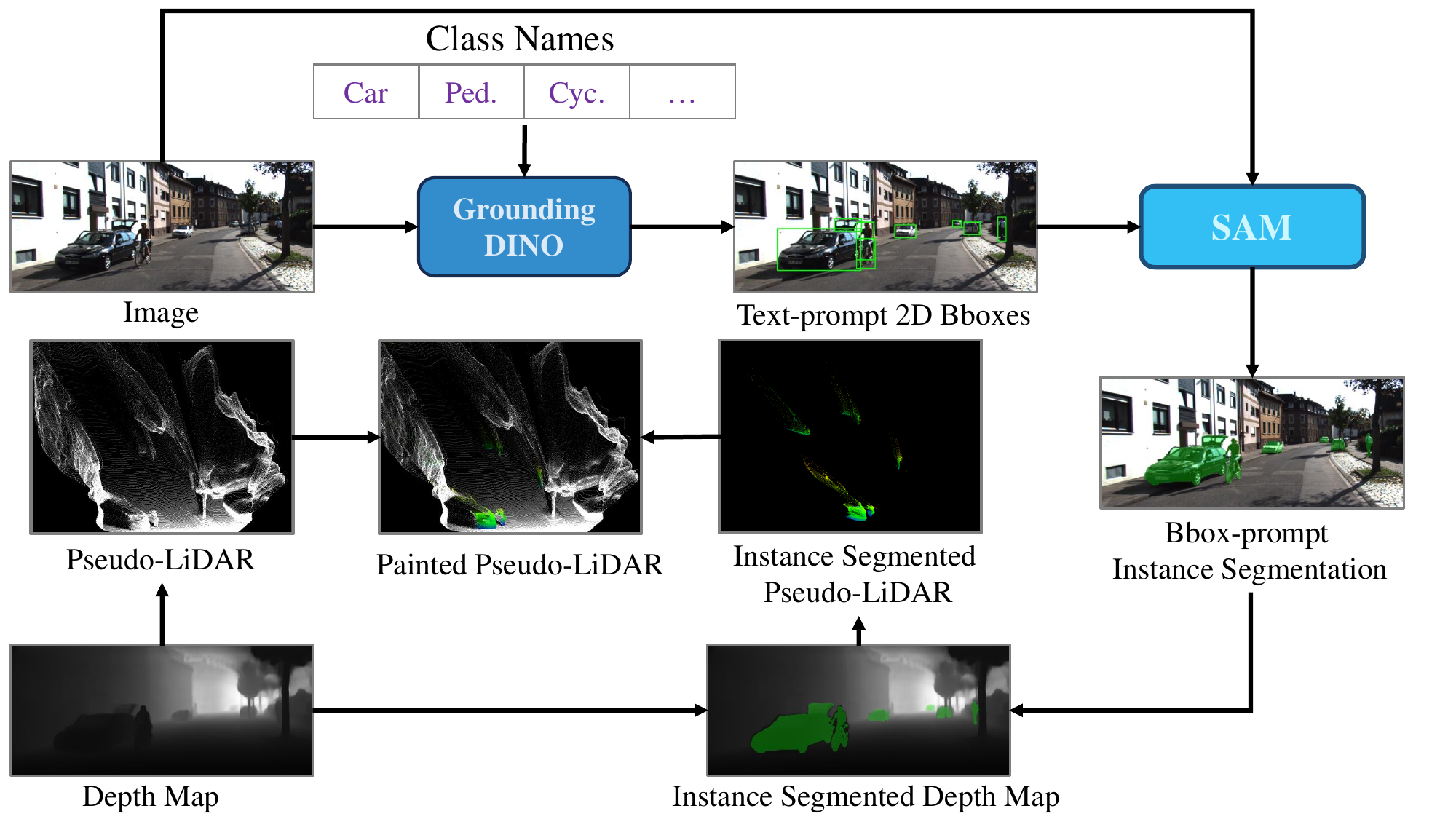}
	\caption{The detailed architecture of Pseudo-LiDAR Painting Text-Prompt SAM.}
	\label{fig:sampanitng}
\end{figure}

\textbf{Preliminaries.}
The \textbf{Segment Anything Model (SAM)}~\cite{SAM} represents a significant leap in the field of computer vision as a VFM, trained on an extensive SA-1B dataset, which comprises over 11 million images and an astounding 1 billion masks, endowing it with remarkable generalization capabilities across different scenes and object types. The design of SAM is underpinned by a lightweight mask decoder and a powerful image encoder, which work in tandem to process prompts and produce high-quality segmentation outputs. Its promptable nature allows for flexibility in accepting various forms of input, including points, bounding boxes, and even free-form text, making it accessible for a multitude of applications. 
The SAM family methods~\cite{FastSAM} further expand on this foundation, exploring different configurations and adaptations that enhance the core capabilities of SAM. These methods have been evaluated extensively, showcasing impressive zero-shot performance and the ability to compete with, or even surpass, fully supervised models in various segmentation tasks. 
SAM and its family methods cannot be directly applied to 3D scenes due to the inherent gap between 2D and 3D spatial representations. However, applying them to monocular 3D object detection can naturally provide robust and powerful semantic-level image features for this task.

\noindent\textbf{Depth Map Segmentation by Text-prompt SAM.} 
As described in  PointPainting~\cite{Pointpainting} and MonoPseudo~\cite{MonoPseudo}, 
the semantic information extracted from the foreground objects in an image through semantic segmentation plays a crucial role in enhancing the accuracy and robustness of 3D object detection. To showcase the broad adaptability of our method, we avoid the use of 2D object detection models that have been pre-trained specifically on autonomous driving data sets, as well as any additional prior data. As a result, it is unable to offer SAM the usual types of cues such as points, bounding boxes, or masks. Instead, the SAM used in this approach is based on text-based prompts. Specifically, the text-prompt SAM contains two parts: a text-prompt open-set {2D} object detector \textbf{Grounding DINO}~\cite{GroundingDINO} and a \textbf{SAM} as shown in Fig.~\ref{fig:sampanitng}. First, the Grounding DINO takes RGB image $I$ in and get {2D} object detection BBoxes $B_{2D}=\{(u_B^{(k)},v_B^{(k)},h_B^{(k)},l_B^{(k)})\}^K_{k=1}$ by text-prompt (\ie 
Car, Pedestrian and Cyclist on KITTI dataset.), where $K$ is the number of predicted BBoxes and $u_B,v_B,h_B,l_B$ denote the center pixel coordinates and the size of each predicted BBox, respectively. Then put the RGB image $I$ and predicted {2D} BBoxes $B_{2D}$ into SAM, and set the BBoxes $B_{2D}$ as prompt to get the semantic segmentation masked image $I_M$. Then we project the mask of $I_M$ onto its corresponding depth map $D$ generated from DAM to get the foreground masked depth map $D_M$. As inspired by Pointpainting~\cite{Pointpainting}, we add the RGB-channel of pixels in $D_M$ to its corresponding pseudo-LiDAR $P$ to get the painted pseudo-LiDAR $\hat{P}$. The details of the painting algorithm are shown in Algorithm~\ref{alg:painting}, 
{the $\mathbf{0}_3$ in Algorithm~\ref{alg:painting} is a zero vector in three dimensions, used to align the dimensions between painted and background points.} 

\begin{algorithm}[t]
	\caption{Pseudo-LiDAR Painting with Text-Prompt SAM}\label{alg:painting}
	\begin{algorithmic}
		\Require{~\\
			RGB image $I \in \mathbb{R} ^ {H \times W \times 3}$.\\
			Depth map $D \in \mathbb{R} ^ {H \times W \times 1}$.\\
			Pseudo-LiDAR points $P \in \mathbb{R} ^ {N \times 3}$ projected from $D$.\\
		}
		\Ensure{~\\
			Painted pseudo-LiDAR points $\hat{P} \in \mathbb{R} ^ {N \times 3+3}$
			\\}
	\end{algorithmic}	
	\begin{algorithmic}
		\State $B_{2D} = Grounding DINO(I)$ \Comment{$B_{2D} \in \mathbb{R} ^ {K \times 4}$}
		\State $I_M = SAM(I, B_{2D})$ \Comment{$I_M \in bool ^ {H \times W}$}
		\State $D_M = D \times I_M$ \\
	\end{algorithmic}
	\begin{algorithmic}
		\For {$d \in D_M, p \in P$}
		\If {$d \ne 0$}
		\State $i = I[u,v,:]$ \Comment{$i \in \mathbb{R} ^ 3$}
		\State $\hat{p} =$Concatenate$(p,i)$ \Comment{$\hat{p} \in \mathbb{R} ^ 6$}
		\Else
		\State $\hat{p} =$Concatenate$(p,\mathbf{0}_3)$ \Comment{$\hat{p} \in \mathbb{R} ^ 6$}
		\EndIf
		\EndFor
	\end{algorithmic}
\end{algorithm}

\subsection{Sparsify Pseudo-LiDAR}
\label{subsec:sparsify}

\begin{algorithm}[!ht]
	\caption{Sparsify Painted Pseudo-LiDAR}
	\label{alg:sparsification}
	\begin{algorithmic}
		\Require{Dense pseudo-LiDAR points $P \in \mathbb{R}^{N \times 6}$}
		\Ensure{Sparse painted pseudo-LiDAR $\bar{P} \in \mathbb{R}^{\bar{N} \times 6}$}\\
	\end{algorithmic}
	
	\noindent\textbf{Step 1: Convert to Spherical Coordinates}
	\begin{algorithmic}
		
		\For{$\hat{p} \in \hat{P}$}
		\State $r_i = \sqrt{x_i^2 + y_i^2 + z_i^2}$
		\State $\theta_i = \arccos\left(\frac{z_i}{r_i}\right)$
		\State $\phi_i = \arctan2(y_i, x_i)$
		\If{$\theta_i > 1.5$}
		\State $\hat{p_i^s}$ = Concatenate$(\hat{p},r_i,\theta_i,\phi_i)$\Comment{$p_i^s \in \mathbb{R}^9$}
		\Else
		\State Remove $p_i$
		\EndIf
		\EndFor
		\State Obtain $\hat{P^s}=\{p^s_1,p^s_2,...,p^s_{n_s}\}$
		\\
	\end{algorithmic}
	
	\textbf{Step 2: Spherical Voxelization and Sampling}
	\begin{algorithmic}
		\State  $\hat{V} \gets$ Voxelize $\hat{P^s}$ space into size-predefined voxel size.
		\For {voxel $\hat{v} \in \hat{V}$}
		\State $\hat{P_d} \gets$ Retain the point in $\hat{v_i}$ with the smallest $r$.
		\EndFor \\
	\end{algorithmic}
	
	\textbf{Step 3: Filter outliers and Output}
	\begin{algorithmic}
		\State $\hat{P_d^c} \gets$ Filter points in $\hat{P_d}$ outlier pre-defined range.
		\State $\hat{V^c} \gets$ Voxelize $\hat{P_d^c}$ with larger voxel size.
		\State $\bar{P} \gets$ Randomly sample points in $\hat{V^c}$
	\end{algorithmic}
\end{algorithm}

Many recent monocular~\cite{MonoPseudo, Pseudo3d} or multimodal fusion-based~\cite{Pseudo-LiDAR++, Virconv} 3D object detectors utilize pseudo-LiDAR generated from images through depth estimation or completion algorithms. However, pseudo-LiDAR is much denser compared to true LiDAR captured by scans, as it has an equal number of pseudo-points as the number of pixels in the image. Additionally, as mentioned in VirConv~\cite{Virconv}, the original pseudo-LiDAR introduces a significant computational cost, leading to a decrease in detection speed (more than $2$--$3\times$ slower than true LiDAR-based 3D object detectors). Furthermore, too many background points may distract the networks from the foreground objects since only the points corresponding to the foreground objects are the focus in LiDAR-based 3D object detectors. 
Besides the issues mentioned above, pseudo-LiDAR also suffers from noise problems. These noises stem from inaccuracies in depth estimation, resulting in non-Gaussian distributed noise. Traditional methods struggle to eliminate such noise. In Pseudo-LiDAR++~\cite{Pseudo-LiDAR++}, cheap 2-beams true LiDAR point clouds are used for pseudo-LiDAR correction to mitigate these noises inexpensively. However, our proposed method relies solely on single-image data. Therefore, we employ some traditional methods to minimize the impact of these noises as much as possible. 
{The sparsify operation is shown in Fig.~\ref{fig:overview} and Algorithm~\ref{alg:sparsification}.}
Specifically, after getting the painted pseudo-LiDAR $\hat{P}$, we first map each point in $\hat{P}$ to spherical coordinate space and voxelize the $\hat{P}$ with a small voxel size using the coordinates in this spherical space. We retain only the mean {3D} coordinates of all points within each voxel to get the denoised painted pseudo-LiDAR $\hat{P}_d$.
Then we filter out the outliers pseudo points in $\hat{P}_d$ based on predefined ranges. The specific filtering ranges will be set differently according to different datasets, which will be detailed in Section~\ref{sec:exp}.
Finally, we voxelize based on the 3D coordinates in the pseudo-LiDAR using a larger voxel size, and then randomly sample points within each voxel to be less than or equal to a fixed number of points (5 in this method), we can get the final sparse painted pseudo-LiDAR $\bar{P} \in \mathbb{R} ^ {\bar{N} \times 6}, \bar{N} \ll N $.

\subsection{LiDAR-Based Detection}
\label{subsec:lidardetection}

The painted pseudo-LiDAR $\bar{P}$ can be input into any LiDAR-based 3D object detector to obtain 3D object results. In this work, we focus on LiDAR-only 3D detectors, although our method allows for the use of any multimodal fusion (i.e., inputting LiDAR + image) based 3D detector. This decision is made for the sake of detection speed, considering that the pseudo-LiDAR point count is significantly higher compared to real point counts. Employing a multimodal fusion-based detector would substantially decrease detection speed. Therefore, we only consider single LiDAR-based 3D detectors in this work.

Specifically, we show that VFMM3D is compatible with three different LiDAR-based detectors: PV-RCNN~\cite{PVRCNN}, Voxel-RCNN~\cite{VoxelRCNN} and PointPillars~\cite{Pointpillars}.
These include widely-used LiDAR detectors, each featuring a distinct network architecture: single-stage (PointPillars) versus two-stage (PV-RCNN and Voxel-RCNN). \\
\noindent\textbf{Voxel-RCNN.}
The Voxel-RCNN method is designed to enhance the performance of voxel-based 3D object detection while maintaining computational efficiency. It introduces a two-stage framework that includes a 3D backbone network for feature extraction, a 2D bird-eye-view (BEV) Region Proposal Network (RPN), and a detection head. The key innovation is voxel RoI pooling, which straightforwardly extracts RoI features from voxel features for subsequent refinement. This approach allows for real-time frame processing while achieving comparable detection accuracy to that of state-of-the-art point-based approaches, significantly reducing computation costs. 

\noindent\textbf{PV-RCNN.}
PV-RCNN (Point-Voxel Feature Set Abstraction for 3D Object Detection) is a high-performance 3D object detection framework that deeply integrates 3D voxel Convolutional Neural Networks with PointNet-based set abstraction. It leverages the efficient learning of the 3D voxel CNN for high-quality proposals and the flexible receptive fields of PointNet-based networks for capturing accurate location information and context. The framework summarizes the 3D scene into a small set of keypoints via a novel voxel set abstraction module and then aggregates these keypoint features to RoI-grid points for proposal refinement. This integration of point-based and voxel-based feature learning leads to improved performance in 3D object detection with manageable memory consumption.

\noindent\textbf{PointPillars.}
PointPillars is a method designed for efficient object detection from point clouds, particularly for autonomous driving applications.
It introduces a novel encoder that employs PointNets to learn representations of point clouds organized into vertical columns or pillars.
The encoded features can be integrated with any standard 2D convolutional detection architecture, and PointPillars introduces a streamlined downstream network for this integration.
By using pillars rather than voxels, PointPillars eliminates the need for hand-tuning the binning of the vertical direction and leverages the efficiency of 2D convolutions on a GPU. The method significantly outperforms previous encoders in both speed and accuracy, running at 62 Hz and setting new standards for performance on KITTI benchmarks for both 3D and bird's eye view detection.

\section{Experiments}
\label{sec:exp}

\subsection{Settings}
\noindent\textbf{Datasets.} We evaluate our method on two datasets, \textbf{Waymo} and \textbf{KITTI}.
The Waymo dataset~\cite{waymo} contains 1,150 video sequences collected from diverse driving environments. We use the official splitting protocol to split the dataset into a train set with 798 sequences, 158081 samples and a validation set with 202 sequences, 39,848 samples. We follow DEVIANT~\cite{deviant}, PCT~\cite{PCT} and other methods~\cite{MonoNeRD, MonoJSG}, using the front view camera for monocular 3D object detection. The KITTI dataset contains two benchmarks, 3D object detection and bird’s eye viewk~\cite{KITTI}, which comprises 7,481 training samples and 7,518 testing samples, along with the corresponding LiDAR point clouds, stereo images, and full camera matrix. We divided the original training samples into a \textit{train} spilt set with 3,712 samples and a \textit{validation} split set with 3,769 samples following the previous methods~\cite{ MonoPseudo}. It is worth noting that during both training and testing phases, our approach exclusively utilizes single-view RGB image data and does not incorporate any LiDAR point cloud or stereo image data.

\noindent\textbf{Evaluation Metrics.}
On Waymo, we follow the official evaluation metrics, evaluate on two object levels: Level 1 and Level 2 with $mAP$ and $mAPH$. It assigns each object to a level according to the number of LiDAR points contained within its 3D bounding box. Besides, we also provide the performance on three distance ranges: [0m, 30m), [30m, 50m), [50m, $\infty$].
For KITTI, we report the performance of our method with the standard evaluation metric $AP_{40}$,
in terms of average precision sampled at 40 recall positions in the precision-recall curve, under three different difficulties (easy, moderate, and hard). 
For broad validation, we selected two IoU thresholds for car class on 3D boxes and BEV boxes, namely 0.7 ($AP@0.7$) and 0.5 ($AP@0.5$).

\begin{table*}[!t]
    \centering
    \caption{Comparison Result on Waymo \textit{val.} set for the vehicles class with \textbf{IoU=0.7}. \textcolor{red}{Red} indicates the best results, while \textcolor{blue}{blue} indicates the second-best results.}
    \label{tab:waymo_07}
    \fontsize{8}{10}\selectfont
        \begin{tabular}{llccccc}
            \toprule
            \multirow{2}{*}{Diff.} & \multirow{2}{*}{Extra} & \multirow{2}{*}{Method} & \multicolumn{4}{c}{3D mAP / mAPH (IoU=0.7)} \\
            \cmidrule(lr){4-7}
            & & & Overall & 0 - 30m & 30 - 50m & 50m - $\infty$ \\
            \midrule
            
            \multirow{13}{*}{L1} 
            & LiDAR & CaDDN~\cite{CaDNN} & 5.03 / 4.99 & 14.54 / 14.43 & 1.47 / 1.45 & 0.10 / 0.10 \\
            & LiDAR & MonoNeRD~\cite{MonoNeRD} & 10.66 / 10.56 & 27.84 / 27.57 & 5.40 / 5.36 & 0.72 / 0.71 \\
            & LiDAR & DID-M3D~\cite{did-m3d} & - / - & - / - & - / - & - / - \\
            \cmidrule{2-7}
            & Depth & PatchNet~\cite{PatchNet} & 0.39 / 0.37 & 1.67 / 1.63 & 0.13 / 0.12 & 0.03 / 0.03 \\
            & Depth & PCT~\cite{PCT} & 0.89 / 0.88 & 3.18 / 3.15 & 0.27 / 0.27 & 0.07 / 0.07 \\
            \cmidrule{2-7}
            & - & M3D-RPN~\cite{M3D-RPN} & 0.35 / 0.34 & 1.12 / 1.10 & 0.18 / 0.18 & 0.02 / 0.02 \\
            & - & GUPNet~\cite{GUPNet} & 2.28 / 2.27 & 6.15 / 6.11 & 0.81 / 0.80 & 0.03 / 0.03 \\
            & - & DEVIANT~\cite{deviant} & 2.69 / 2.67 & 6.95 / 6.90 & 0.99 / 0.98 & 0.02 / 0.02 \\
            & - & MonoJSG~\cite{MonoJSG} & 0.97 / 0.95 & 4.65 / 4.59 & 0.55 / 0.53 & 0.10 / 0.09 \\
            & - & SSD-MonoDETR~\cite{Ssd-monodetr} & 4.54 / - & 9.93 / - & 1.18 / - & 0.15 / - \\
            & - & MonoSIM~\cite{monosim} & 1.60 / 1.59 & 6.08 / 6.02 & 0.40 / 0.39 & 0.01 / 0.01 \\
            & - & VFMM3D (Voxel) & \textcolor{blue}{8.57} / \textcolor{blue}{8.34} & \textcolor{blue}{27.81} / \textcolor{blue}{27.21} & \textcolor{blue}{2.39} / \textcolor{blue}{2.24} & \textcolor{red}{0.28} / \textcolor{red}{0.23} \\
            & - & VFMM3D (PV) & \textcolor{red}{8.97} / \textcolor{red}{8.72} & \textcolor{red}{29.55} / \textcolor{red}{28.90} & \textcolor{red}{2.69} / \textcolor{red}{2.49} & 0.13 / 0.10 \\
            
            \midrule
            \midrule
            
            \multirow{13}{*}{L2} 
            & LiDAR & CaDDN & 4.49 / 4.45 & 14.50 / 14.38 & 1.42 / 1.41 & 0.09 / 0.09 \\
            & LiDAR & MonoNeRD~\cite{MonoNeRD} & 10.03 / 9.93 & 27.75 / 27.48 & 5.25 / 5.21 & 0.60 / 0.59 \\
            & LiDAR & DID-M3D~\cite{did-m3d} & - / - & - / - & - / - & - / - \\
            \cmidrule{2-7}
            & Depth & PatchNet~\cite{PatchNet} & 0.38 / 0.36 & 1.67 / 1.63 & 0.13 / 0.11 & 0.03 / 0.03 \\
            & Depth & PCT~\cite{PCT} & 0.66 / 0.66 & 3.18 / 3.15 & 0.27 / 0.26 & 0.07 / 0.07 \\
            \cmidrule{2-7}
            & - & M3D-RPN~\cite{M3D-RPN} & 0.33 / 0.33 & 1.12 / 1.10 & 0.18 / 0.17 & 0.02 / 0.02 \\
            & - & GUPNet~\cite{GUPNet} & 2.14 / 2.12 & 6.13 / 6.08 & 0.78 / 0.77 & 0.02 / 0.02 \\
            & - & DEVIANT~\cite{deviant} & 2.52 / 2.50 & 6.93 / 6.87 & 0.95 / 0.94 & 0.02 / 0.02 \\
            & - & MonoJSG~\cite{MonoJSG} & 0.91 / 0.89 & 4.64 / 4.65 & 0.55 / 0.53 & 0.09 / 0.09 \\
            & - & SSD-MonoDETR~\cite{Ssd-monodetr} & 4.12 / - & 8.87 / - & 1.02 / - & \textcolor{blue}{0.13} / - \\
            & - & MonoSIM~\cite{monosim} & 1.49 / 1.48 & 6.05 / 6.00 & 0.38 / 0.38 & 0.004 / 0.004 \\
            & - & VFMM3D (Voxel) & \textcolor{blue}{7.50} / \textcolor{blue}{7.30} & \textcolor{blue}{27.33} / \textcolor{blue}{26.75} & \textcolor{blue}{2.18} / \textcolor{blue}{2.05} & \textcolor{red}{0.22} / \textcolor{red}{0.18} \\
            & - & VFMM3D (PV) & \textcolor{red}{7.85} / \textcolor{red}{7.63} & \textcolor{red}{29.06} / \textcolor{red}{28.42} & \textcolor{red}{2.46} / \textcolor{red}{2.27} & 0.11 / 0.08 \\
            
            \bottomrule
        \end{tabular}%
\end{table*}

\begin{table*}[!t]
    \centering
    \caption{Comparison Result on Waymo \textit{val.} set for the vehicles class with \textbf{IoU=0.5}. \textcolor{red}{Red} indicates the best results, while \textcolor{blue}{blue} indicates the second-best results.}
    \label{tab:waymo_05}
    \fontsize{8}{10}\selectfont
        \begin{tabular}{llccccc}
            \toprule
            \multirow{2}{*}{Diff.} & \multirow{2}{*}{Extra} & \multirow{2}{*}{Method} & \multicolumn{4}{c}{3D mAP / mAPH (IoU=0.5)} \\
            \cmidrule(lr){4-7}
            & & & Overall & 0 - 30m & 30 - 50m & 50m - $\infty$ \\
            \midrule
            
            \multirow{13}{*}{L1} 
            & LiDAR & CaDDN~\cite{CaDNN} & 17.54 / 17.31 & 45.00 / 44.46 & 9.24 / 9.11 & 0.64 / 0.62 \\
            & LiDAR & MonoNeRD~\cite{MonoNeRD} & 31.18 / 30.70 & 61.11 / 60.28 & 26.08 / 25.71 & 6.60 / 6.47 \\
            & LiDAR & DID-M3D~\cite{did-m3d} & 20.66 / 20.47 & 40.92 / 40.60 & 15.63 / 15.48 & 5.35 / 5.24 \\
            \cmidrule{2-7}
            & Depth & PatchNet~\cite{PatchNet} & 2.92 / 2.74 & 10.03 / 9.75 & 1.09 / 0.96 & 0.23 / 0.18 \\
            & Depth & PCT~\cite{PCT} & 4.20 / 4.15 & 14.70 / 14.54 & 1.78 / 1.75 & 0.39 / 0.39 \\
            \cmidrule{2-7}
            & - & M3D-RPN~\cite{M3D-RPN} & 3.79 / 3.63 & 11.14 / 10.70 & 2.16 / 2.09 & 0.26 / 0.21 \\
            & - & GUPNet~\cite{GUPNet} & 10.02 / 9.94 & 24.78 / 24.59 & 4.84 / 4.78 & 0.22 / 0.22 \\
            & - & DEVIANT~\cite{deviant} & 10.98 / 10.89 & 26.85 / 26.64 & 5.13 / 5.08 & 0.18 / 0.18 \\
            & - & MonoJSG~\cite{MonoJSG} & 5.65 / 5.47 & 20.86 / 20.26 & 3.91 / 3.79 & 0.97 / 0.92 \\
            & - & SSD-MonoDETR~\cite{Ssd-monodetr} & 11.83 / - & 27.69 / - & 5.33 / - & 0.85 / - \\
            & - & MonoSIM~\cite{monosim} & 8.16 / 8.04 & 25.86 / 25.48 & 3.62 / 3.55 & 0.08 / 0.08 \\
            & - & VFMM3D (Voxel) & \textcolor{blue}{19.85} / \textcolor{blue}{19.01} & \textcolor{blue}{56.21} / \textcolor{blue}{54.40} & \textcolor{blue}{9.51} / \textcolor{blue}{8.79} & \textcolor{red}{1.66} / \textcolor{red}{1.30} \\
            & - & VFMM3D (PV) & \textcolor{red}{20.20} / \textcolor{red}{19.34} & \textcolor{red}{58.11} / \textcolor{red}{56.36} & \textcolor{red}{10.57} / \textcolor{red}{9.61} & \textcolor{blue}{0.98} / 0.71 \\
            
            \midrule
            \midrule
            
            \multirow{13}{*}{L2} 
            & LiDAR & CaDDN & 16.51 / 16.28 & 44.87 / 44.33 & 8.99 / 8.86 & 0.58 / 0.55 \\
            & LiDAR & MonoNeRD~\cite{MonoNeRD} & 29.29 / 28.84 & 60.91 / 60.08 & 25.36 / 25.00 & 5.77 / 5.66 \\
            & LiDAR & DID-M3D~\cite{did-m3d} & 19.37 / 19.19 & 40.77 / 40.46 & 15.18 / 15.04 & 4.69 / 4.59 \\
            \cmidrule{2-7}
            & Depth & PatchNet~\cite{PatchNet} & 2.42 / 2.28 & 10.01 / 9.73 & 1.07 / 0.94 & 0.22 / 0.16 \\
            & Depth & PCT~\cite{PCT} & 4.03 / 3.99 & 14.67 / 14.51 & 1.74 / 1.71 & 0.36 / 0.35 \\
            \cmidrule{2-7}
            & - & M3D-RPN~\cite{M3D-RPN} & 3.61 / 3.46 & 11.12 / 10.67 & 2.12 / 2.04 & 0.24 / 0.20 \\
            & - & GUPNet~\cite{GUPNet} & 9.39 / 9.31 & 24.69 / 24.50 & 4.67 / 4.62 & 0.19 / 0.19 \\
            & - & DEVIANT~\cite{deviant} & 10.29 / 10.20 & 26.75 / 26.54 & 4.95 / 4.90 & 0.16 / 0.16 \\
            & - & MonoJSG~\cite{MonoJSG} & 5.34 / 5.17 & 20.79 / 20.19 & 3.79 / 3.67 & 0.85 / 0.82 \\
            & - & SSD-MonoDETR~\cite{Ssd-monodetr} & 11.34 / - & 27.62 / - & 5.21 / - & 0.76 / - \\
            & - & MonoSIM~\cite{monosim} & 7.58 / 7.47 & 25.75 / 25.37 & 3.49 / 3.43 & 0.07 / 0.07 \\
            & - & VFMM3D (Voxel) & \textcolor{blue}{17.42} / \textcolor{blue}{16.68} & \textcolor{blue}{55.39} / \textcolor{blue}{53.60} & \textcolor{blue}{8.72} / \textcolor{blue}{8.05} & \textcolor{red}{1.32} / \textcolor{red}{1.03} \\
            & - & VFMM3D (PV) & \textcolor{red}{17.73} / \textcolor{red}{16.97} & \textcolor{red}{57.30} / \textcolor{red}{55.58} & \textcolor{red}{9.70} / \textcolor{red}{8.81} & 0.78 / 0.56 \\
            
            \bottomrule
        \end{tabular}%
\end{table*}

\noindent\textbf{Training Details.}
The size of images is set to 1224$\times$370 pixels on the KITTI dataset and 1920$\times$ 1280 pixels on the Waymo dataset without any 2D data augmentation. In the LiDAR-based detection stage, we use 3D random horizontal flip augmentation and point-shuffle augmentation. Due to the inaccuracy of depth estimation, there is a significant deviation between the pseudo points inside the 3D annotated boxes and the real points. Therefore, in this method, we do not utilize the data augmentation method of GT-sampling~\cite{SECOND}. We use Voxel-RCNN~\cite{VoxelRCNN} as the LiDAR-based 3D object detector for experiments on the KITTI dataset. For Voxel-RCNN and PV-RCNN~\cite{PVRCNN}, we set the pseudo-LiDAR range as [0, 70.4], [-40.0, 40.0], [-3.0, 1.0] meters along the X, Y, and Z axes, respectively. And the voxel size is (0.05, 0.05, 0.1) meters. For PointPillars~\cite{Pointpillars}, we set the pseudo-LiDAR range as [0, 69.12], [-39.68, 39.68], [-3.0, 1.0] meters, the pillar size is set as (0.16, 0.16, 4) meters. On the Waymo dataset, we use Voxel-RCNN and PV-RCNN and set the pseudo-LiDAR range as [0, 75.2], [-75.2, 75.2], [-2.0, 4.0] meters with voxel size (0.1, 0.1, 0.15) meters and [0, 59.6], [-25.6, 25.6], [-2.0, 4.0] meters with voxel size (0.05, 0.05, 0.15) meters, respectively.

Our method is primarily implemented using OpenPCDet~\cite{openpcdet}. We utilize pre-trained models, including Grounding DINO, SAM with vit-h, and DAM with vit-l, which are fine-tuned on KITTI outdoor metric depth datasets to contain metric depth estimation ability.
However, we note that the officially provided DAM model is finetuned on the whole KITTI-depth dataset, as mentioned in DD3D~\cite{DD3D} and Pesuo-Lidar++~\cite{Pseudo-LiDAR++}. This training set overlaps with the KITTI-3D validation data for detection.
Thus, we use the KITTI-3D clean training split, which removes training images that are geographically close to any of the KITTI-3D images provided by DD3D to finetune the DAM to avoid any bias.
We also demonstrate the results on KITTI with DAM finetuned with the whole KITTI-depth dataset as some methods~\cite{DDMP-3D} also report this kind of results, as shown in Table~\ref{tab:kitti} with the indicator *.

The LiDAR-based 3D object detectors are trained on the sparse painted pseudo-LiDAR using 8 RTX A6000 GPUs. We employ the AdamW~\cite{AdamW} optimizer with a learning rate of 4e-4 and weight decay of 0.01. The batch size is set to 32 for PointPillars, 32 for Voxel-RCNN, and 8 for PV-RCNN. Additionally, the number of training epochs is set to 80 for all detectors.

\begin{table*}[!t]
	\centering
	\caption{Comparison of our model with state-of-the-art models on the KITTI \textit{val.} set for the car class. 'Mod.' indicates the moderate difficulty level. * indicates that DAM is fine-tuned on the whole KITTI-depth set.
    \textcolor{red}{Red} indicates the best results, while \textcolor{blue}{blue} indicates the second-best results.}
	\label{tab:kitti}
	\fontsize{8}{10}\selectfont
    \setlength{\tabcolsep}{2pt}
	\begin{tabular}{lcccccccccccc}
		\toprule
		\multirow{2}{*}{Method}
		& \multicolumn{3}{c}{3D AP@0.7}
		& \multicolumn{3}{c}{BEV AP@0.7} & \multicolumn{3}{c}{3D AP@0.5} & \multicolumn{3}{c}{BEV AP@0.5} \\
		\cmidrule(lr){2-4} \cmidrule(lr){5-7}\cmidrule(lr){8-10}\cmidrule(lr){11-13}
		& Easy & Mod. & Hard & Easy & Mod. & Hard & Easy & Mod. & Hard & Easy & Mod. & Hard \\
		\midrule
		Monopair~\cite{MonoPair}
		& 16.28 & 12.30 & 10.42 & 24.12 & 18.17 & 15.76 & 55.38 & 42.39 & 37.99 & 61.06 & 47.63 & 41.92 \\
		MonoDLE~\cite{MonoDLE}
		& 17.45 & 13.66 & 11.68 & 24.97 & 19.33 & 17.01& 55.41 & 43.42 & 37.81 & 60.73 & 46.87 & 41.89 \\
		MonoFlex~\cite{MonoFlex}
		& 24.22 & 17.34 & 15.13 & 31.65 & 23.29 & 20.02 & 60.70 & 45.65 & 39.91 & 66.26 & 49.30 & 44.42 \\
		GUPNet~\cite{GUPNet}
		& 22.76 & 16.46 & 13.72 & 31.07 & 22.94 & 19.75 & 57.62 & 42.33 & 37.59 & 61.78 & 47.06 & 40.88 \\
		DDMP-3D*~\cite{DDMP-3D}
		& 28.12 & 20.39 & 16.34 & - & - & - & - & - & - & - & - & - \\
		CaDNN~\cite{CaDNN}
		& 23.57 & 16.31 & 13.84 & - & - & -& - & - & - & - & - & - \\
		MonoRUn~\cite{MonoRUn}
		& 20.02 & 14.65 & 12.61 & - & - & - & 59.71 & 43.39 & 38.44 & - & - & - \\
		MonoDTR~\cite{MonoDTR}
		& 24.52 & 18.57 & 15.51 & 33.33 & 25.35 & 21.68& 64.03 & 47.32 & 42.20 & 69.04 & 52.47 & 45.90 \\
		MonoGround~\cite{MonoGround}
		& 25.24 & 18.69 & 15.58 & 32.68 & 24.79 & 20.56 & 62.60 & 47.85 & 41.97 & 67.36 & 51.83 & 45.65 \\
		MonoPoly-L~\cite{MonoPoly}
		& 22.94 & 16.13 & 13.01 & 29.57 & 22.23 & 19.75 & - & - & -& - & - & -  \\
		MonoNeRD~\cite{MonoNeRD}
		& 20.64 & 15.44 & 13.99 & 29.03 & 22.03 & 19.41 & - & - & -& - & - & -  \\
		PDR~\cite{PDR}
		& 27.65 & 19.44 & 16.24 & 35.59 & 25.72 & 21.35 & - & - & -& - & - & -  \\
		MonoSGC~\cite{MonoSGC}
		& 28.76 & 19.55 & 16.68 & - & - & - & - & - & -& - & - & -  \\
		MonoOAPC~\cite{MonoOAPC}
		& 23.91 & 16.95 & 14.13 & - & - & - & - & - & -& - & - & -  \\
		UniCuboid~\cite{Unicuboid}
		& 26.87 & 20.10 & 17.13 & - & - & - & - & - & -& - & - & - \\
		\midrule
		VFMM3D (PointPillars)
		& 22.43 & 15.45 & 13.92 & 35.85 & 24.88 & 22.75 & 63.52 & 45.60 & 41.79 & 68.26 & 48.62 & 44.75 \\
		VFMM3D (Voxel-RCNN)
		& 29.09 & 19.41 & 17.09 & 39.43 & 26.56 & 23.69& 68.72 & 50.53 & 45.02 & 73.17 & 52.90 & 47.25 \\
		VFMM3D (PV-RCNN)
		& 29.05 & 19.10 & 16.86 & 41.78 & 28.53
		& \textcolor{red}{25.61} & 69.95 & 51.78 & \textcolor{red}{47.01} & 74.24 & 54.98 & \textcolor{blue}{49.69} \\
		\midrule
		VFMM3D (PointPillars)*
		& 27.23 & 17.15 & 13.95 & 38.02 & 24.93 & 20.79 & 69.94 & 48.75 & 42.98 & 74.12 & 54.37 & 48.09 \\
		VFMM3D (Voxel-RCNN)*
		& \textcolor{red}{34.60}
		& \textcolor{red}{21.58}
		& \textcolor{red}{18.23}
		& \textcolor{red}{44.18}
		& \textcolor{blue}{28.66}
		& 24.02 & \textcolor{blue}{72.85} & \textcolor{blue}{51.83} & 45.53 & \textcolor{blue}{76.69} & \textcolor{blue}{55.43} & 48.95 \\
		VFMM3D (PV-RCNN)*
		& \textcolor{blue}{32.06}
		& \textcolor{blue}{21.00}
		& \textcolor{blue}{17.49}
		& \textcolor{blue}{43.30}
		& \textcolor{red}{28.88}
		& \textcolor{blue}{24.72} & \textcolor{red}{73.21} & \textcolor{red}{53.51} & \textcolor{blue}{46.88} & \textcolor{red}{77.09} & \textcolor{red}{58.37} & \textcolor{red}{51.69} \\
		\bottomrule
	\end{tabular}
\end{table*}

\subsection{Waymo Results}
Tab.~\ref{tab:waymo_07} and Tab.~\ref{tab:waymo_05} shows the performance comparison results on Waymo validation spilt. It demonstrates our method VFMM3D outperforms other monocular methods without extra modality data during training, including depth~\cite{PatchNet,PCT} or sparse lidar point cloud~\cite{CaDNN, MonoNeRD}, which surpasses the latest monocular method DEVIANT~\cite{deviant} by 7.41\% and 6.20\% on level 1 and level 2 3D mAP(IoU=0.5), respectively. It is worth noting that our method even outperforms CaDNN~\cite{CaDNN}, which uses LiDAR as extra training data, 0.72\% on level 1 3D mAP(IoU=0.5).

\subsection{KITTI Results}
\begin{figure*}[!htb]
	\centering
	\includegraphics[width=\textwidth]{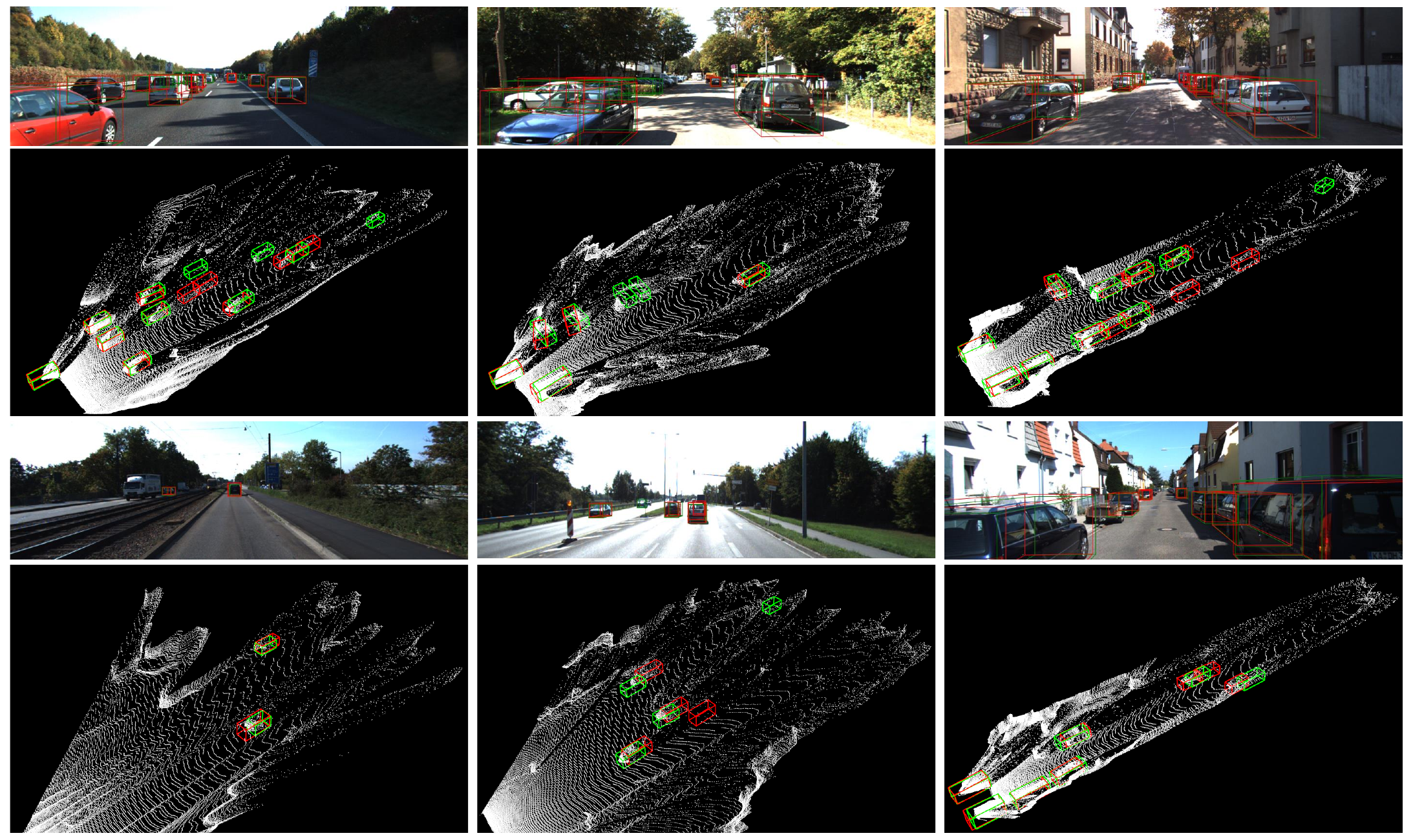}
	\caption{Qualitative results of VFMM3D on KITTI \textit{val.} set. We visualize our 3D bounding box estimates (in \textcolor{red}{red}) alongside ground truth annotations (in {green}) on front view images (1st and 3rd rows) and pseudo-LiDAR point clouds (2nd and 4th rows).}
	\label{fig:qualit}
\end{figure*}

We compare the performance of our method on the KITTI \textit{val.} set for the car class with the state-of-the-art methods. The related results are shown in Tab.~\ref{tab:kitti}. It is evident that our method outperforms existing methods in both 3D and BEV under all difficulty levels. Specifically, when IoU is set to 0.5, our method surpasses these methods by even greater margins. Compared to the latest monocular object detection method MonoNeRD~\cite{MonoNeRD}, our approach surpasses it by 8.41\% and 12.75\% in 3D and BEV at the easy level, respectively. Some qualitative results of 3D detection boxes from our method on KITTI \textit{val.} set are visualized in Fig.~\ref {fig:qualit}.

\begin{figure*}[!htb]
	\centering
	\includegraphics[width=\textwidth]{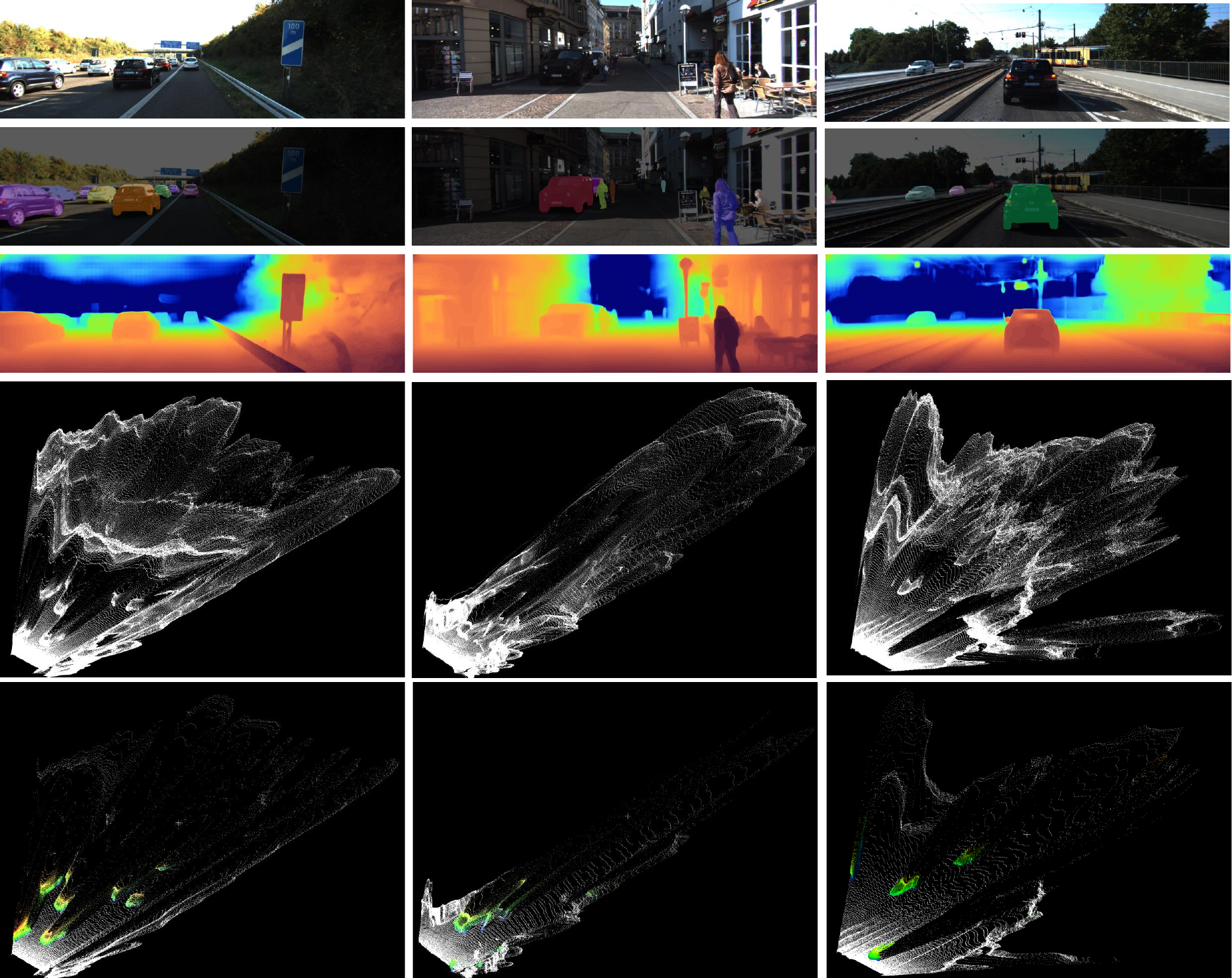}
	\caption{Visualization of segmentation results (2nd row) from text-prompt SAM, depth maps (3rd row) from DAM, raw Pseudo-LiDAR (4th row) from depth map and Painted Sparse Pseudo-LiDAR (5th row) results from each component of VFMM3D on KITTI.}
	\label{fig:vis}
\end{figure*}

\subsection{Visualization of Each Component Results.}
We visualize the results of each component in VFMM3D in Fig.~\ref{fig:vis}. In the second and third rows, both Text-prompt SAM and DAM produce good segmentation results and accurate depth maps. Comparing the fourth and fifth rows, it is evident that our sparse pseudo-LiDAR obtained by our designed sparsify and painting operations can reduce a significant amount of useless noise points when enhancing information for foreground points, as opposed to the dense pseudo-LiDAR directly derived from the depth map.

\begin{table}[!t]
    \centering
    \caption{Ablation analysis on KITTI \textit{val.} set. We quantify the impacts of painting strategy, \textit{RGB P} means painting all pseudo-LiDAR with corresponding RGB pixel, \textit{SAM P} refers to pseudo-LiDAR painting by text-prompt SAM.}
    \label{tab:abliation}
    \fontsize{8}{10}\selectfont
    \resizebox{\linewidth}{!}{
        \begin{tabular}{ccccccccc}
            \toprule
            \multirow{2}{*}{DAM} & \multirow{2}{*}{RGB-P} & \multirow{2}{*}{SAM-P} & \multicolumn{3}{c}{3D AP@0.7} & \multicolumn{3}{c}{BEV AP@0.7} \\
            \cmidrule(lr){4-6} \cmidrule(lr){7-9}
            & & & Easy & Mod. & Hard & Easy & Mod. & Hard \\
            \midrule
            \checkmark & & & 21.50 & 15.88 & 14.05 & 34.81 & 25.38 & 22.79 \\
            \checkmark & \checkmark & & 26.96 & 17.77 & 15.77 & 36.63 & 25.60 & 23.51 \\
            \checkmark & & \checkmark & 29.05 & 19.10 & 16.86 & 41.78 & 28.53 & 25.61 \\
            \bottomrule
        \end{tabular}}
\end{table}

\begin{table*}[!t]
    \centering
    \caption{Ablation analysis on KITTI \textit{val.} set with Voxel-RCNN 3D detector. We compared inference time(IT) and detection results between using different voxel sizes in voxelization and different sparsification ratio(SR) during our pseudo-LiDAR sparsification \textit{step 2} and other sparsification methods.}
    \label{tab:sparsification}
    \fontsize{8}{10}\selectfont
    \setlength{\tabcolsep}{3pt}
        \begin{tabular}{lccccccccc}
            \toprule
            \multirow{2}{*}{Methods} & \multirow{2}{*}{Voxel size} & \multirow{2}{*}{IT(ms)} & \multirow{2}{*}{SR (\%)} & \multicolumn{3}{c}{3D AP@0.5} & \multicolumn{3}{c}{BEV AP@0.5} \\
            \cmidrule(lr){5-7} \cmidrule(lr){8-10}
            & & & & Easy & Mod. & Hard & Easy & Mod. & Hard \\
            \midrule
            No sparsification & - & 91.73 & 0 & 69.88 & 49.64 & 43.40 & 73.57 & 54.20 & 47.89 \\
            \midrule
            Random & - & 49.04 & 90 & 70.77 & 49.53 & 43.30 & 74.58 & 53.21 & 46.79 \\
            \midrule
            Uniform grid & - & 50.14 & 90 & 68.66 & 48.80 & 42.52 & 72.67 & 52.44 & 45.96 \\
            \midrule
            \multirow{3}{*}{Ours} & {[}0.001, 0.001{]} & 65.44 & 5 & 69.81 & 49.65 & 43.43 & 73.45 & 54.10 & 47.84 \\
             & {[}0.002, 0.002{]} & 53.10 & 67 & \textbf{72.85} & \textbf{51.83} & \textbf{45.53} & \textbf{76.69} & \textbf{55.43} & \textbf{48.95} \\
             & {[}0.0025, 0.0025{]} & 50.97 & 80 & 69.48 & 49.08 & 43.09 & 73.53 & 52.70 & 46.44 \\
            \bottomrule
        \end{tabular}%
\end{table*}

\subsection{Ablation Study}
\label{subsec:abl}
In this section, we delve into our method from two key perspectives: the impacts of the pseudo-LiDAR painting strategy and the ablation study of different LiDAR-based 3D object detectors. 

\noindent\textbf{Impacts of Different Painting Strategy.} 
We conduct ablation experiments to assess the impacts of different pseudo-LiDAR painting strategies on the final 3D detection results with the PV-RCNN 3D detector. 
As depicted in Tab.~\ref{tab:abliation}, we observe that the detailed and accurate foreground information provided by the text-prompt SAM (\textit{SAM-P}) enables subsequent LiDAR-based 3D detectors to concentrate on the features of foreground objects efficiently. In comparison to the global painting approach (\textit{RGB-P}), this enhances the overall average detection accuracy by 1.51\% and 3.39\% for 3D and BEV detection results, respectively.

\noindent\textbf{Ablation Study on Different LiDAR-Based 3D Object Detectors.}
We employ various LiDAR-based 3D object detectors, including PointPillars~\cite{Pointpillars}, PV-RCNN~\cite{PVRCNN}, and Voxel-RCNN~\cite{VoxelRCNN}, to showcase the versatility of VFMM3D. The impacts of different 3D detectors on prediction accuracy are illustrated in Tab.~\ref{tab:waymo_05}, Tab.~\ref{tab:waymo_07}, and Tab.~\ref{tab:kitti}.

{
	\noindent\textbf{Ablation Study on Pseudo-LiDAR Sparsification Strategy.}
	As shown in Tab.~\ref{tab:sparsification}, we conduct an ablation study to evaluate the impact of different voxel sizes used in Step 2 of our pseudo-LiDAR sparsification strategy on both the 3D detection performance and inference time on KITTI validation dataset with Voxel-RCNN 3D detetcor. The voxel size directly controls the sparsification ratio, which significantly affects the resulting point cloud density.
	Since the pseudo-LiDAR generated from the depth map contains only one point along the \textit{X}-axis, we fix the voxel size in that direction to 200m. We then vary the voxel sizes along the remaining two axes, testing three configurations: [0.001m, 0.001m], [0.002m, 0.002m], and [0.0025m, 0.0025m]. A larger voxel size produces a sparser point cloud, and vice versa.
	Results show that the configuration [200m, 0.002m, 0.002m] achieves the best trade-off between accuracy and speed—it provides strong detection performance (results in Tab.\ref{tab:waymo_07}, Tab.\ref{tab:waymo_05}, and Tab.~\ref{tab:kitti}) while offering over 2× faster inference compared to the non-sparsified version. We also compare our method against commonly used sparsification strategies, such as random sampling and uniform grid filtering, and show superior performance in both accuracy and runtime. These findings demonstrate the effectiveness and practical value of our sparsification design. These findings validate the effectiveness of our sparsification strategy and demonstrate its practical benefit.
	Additionally, we analyze the impact of sparsification on depth quality. We define erroneous points as those whose depth values deviate from the ground truth by more than 1 meter. Prior to sparsification, the pseudo-LiDAR contains 55.81\% such erroneous points. After applying our proposed sparsification method, this ratio is significantly reduced to 27.95\%, quantitatively demonstrating the denoising capability of our sparsification strategy and its effectiveness in filtering unreliable depth predictions.
}

{
	\noindent\textbf{Ablation Study on Inference Time of Each Module.} To better understand the efficiency of VFMM3D pipeline, we report the inference time of each major module in Tab.~\ref{tab:inference_time}. The overall pipeline includes the SAM and DAM foundation models, our proposed sparsification and painting module, and the final 3D detector, which is Voxel R-CNN in our experiments. All timings are measured on a single RTX 3090 GPU. Our sparsification method introduces minimal overhead (52.7 ms), while the bulk of the time comes from the SAM. Overall, the system remains efficient and practical.
}

\begin{table}[t]
    \centering
    \caption{Inference time of each module in VFMM3D pipeline.}
    \label{tab:inference_time}
    \fontsize{8}{10}\selectfont
        \begin{tabular}{lc}
            \toprule
            Module & Inference Time (ms) \\
            \midrule
            Text-prompt SAM & 342.1 \\
            DAM & 195.3 \\
            Sparsification and painting & 52.7 \\
            3D detector (Voxel-RCNN) & 53.1 \\
            \midrule
            \textbf{Total} & \textbf{643.2} \\
            \bottomrule
        \end{tabular}%
\end{table}

\section{Conclusion}
In this paper, we proposed VFMM3D, a vision foundation model-driven framework for foreground-aware pseudo-LiDAR generation in monocular 3D object detection. By exploiting the depth prior from the Depth Anything Model (DAM) and the foreground prior from the Segment Anything Model (SAM), VFMM3D transforms monocular images into structured pseudo-LiDAR representations without requiring task-specific fine-tuning of the employed foundation models. A foreground-aware pseudo-LiDAR painting operation is introduced to enhance object-related structures, while a sparsification strategy is developed to remove redundant points and improve the compatibility with LiDAR-based 3D detectors. Extensive experiments on the KITTI and Waymo datasets demonstrate that VFMM3D achieves superior performance over existing monocular 3D object detection methods. Furthermore, evaluations with multiple LiDAR-based detectors verify the effectiveness and generality of the proposed pseudo-LiDAR generation framework.

\bibliographystyle{IEEEtran}
\bibliography{ref}

\end{document}